\documentclass{article}

\usepackage{microtype}
\usepackage{graphicx}
\usepackage{booktabs} 
\usepackage{soul}
\usepackage{bm}
\usepackage{hyperref}

\usepackage[accepted]{icml2021}

\icmltitlerunning{Adaptive Weighting Scheme for Automatic Data Augmentation}

\usepackage{amsmath,amssymb,amsfonts}
\usepackage{float}
\usepackage{gensymb}
\usepackage{textcomp}
\usepackage{etoolbox}
\usepackage{xspace}

\makeatletter
\DeclareRobustCommand\onedot{\futurelet\@let@token\@onedot}
\def\@onedot{\ifx\@let@token.\else.\null\fi\xspace}

\def\eg{\emph{e.g}\onedot} 
\def\ie{\emph{i.e}\onedot}

\def\wrt{w.r.t\onedot} 
\def\etal{\emph{et al}\onedot}
\makeatother

\usepackage{subcaption}

\begin{document}

\twocolumn[
\icmltitle{Adaptive Weighting Scheme for Automatic Time-Series Data Augmentation}




\begin{icmlauthorlist}
\icmlauthor{Elizabeth Fons}{UoM}
\icmlauthor{Paula Dawson}{AB}
\icmlauthor{Xiao-jun Zeng}{UoM}
\icmlauthor{John Keane}{UoM}
\icmlauthor{Alexandros Iosifidis}{Au}
\end{icmlauthorlist}

\icmlaffiliation{UoM}{Department of Computer Science, University of Manchester, Manchester, UK}
\icmlaffiliation{AB}{AllianceBernstein, London, UK}
\icmlaffiliation{Au}{Department of Electrical and Computer Engineering, Aarhus University, Aarhus, Denmark}

\icmlcorrespondingauthor{Elizabeth Fons}{elizabeth.fons@manchester.ac.uk}


\vskip 0.3in
]



\printAffiliationsAndNotice{}  

\begin{abstract}
Data augmentation methods have been shown to be a fundamental technique to improve generalization in tasks such as image, text and audio classification.
Recently, automated augmentation methods have led to further improvements on image classification and object detection leading to state-of-the-art performances. Nevertheless, little work has been done on time-series data, an area that could greatly benefit from automated data augmentation given the usually limited size of the datasets. We present two sample-adaptive automatic weighting schemes for data augmentation: the first learns to weight the contribution of the augmented samples to the loss, and the second method selects a subset of transformations based on the ranking of the predicted training loss. We validate our proposed methods on a large, noisy financial dataset and on time-series datasets from the UCR archive. On the financial dataset, we show that the methods in combination with a trading strategy lead to improvements in annualized returns of over 50$\%$, and on the time-series data we outperform state-of-the-art models on over half of the datasets, and achieve similar performance in accuracy on the others.
\end{abstract}

\section{Introduction}

Data augmentation is a mainstream approach to reduce over-fitting and improve generalization in neural networks \cite{Goodfellow2016}. It has shown to improve performance on image classification tasks \cite{AlexK2012,Shorten2019,mixup,Simard2003}, on speech recognition \cite{specaugment,dasr2014, dasr2015,dasr2017} and natural language processing \cite{danlp2017, danlp2018}.

Additionally, data augmentation has been proposed to improve time-series classifications tasks \cite{Um2017,guennec2016, Iwana2020, Wen2020, devries2017} and in financial prediction \cite{Teng2020, Zhang2017,dataaugfin}. 
A major challenge with data augmentation is how to search over the space of transformations that can be applied to samples of a class and can generate augmented samples having the characteristics of the same class. This can be prohibitive given the large number of possible transformations and their associated parameters. For example, if we add noise to a sample, what scale of noise should be used? What transformations work best on a certain dataset? 

As a consequence, how to perform {\it learnable} data augmentation is an open question. Ideally, such methods would allow us to select optimal transformations, select an optimal range of transformation parameters and allow assessment of which augmentations are better used in isolation or need to be combined successively.
To be the best of our knowledge there has little or no work on automatic data augmentation for time-series data, much less for financial data. 

In this paper, we propose two novel automatic augmentation policies to apply on time-series data and compare them with RandAugment \cite{randaugment} - which we apply to time-series data for the first time. The proposed policies combine multiple augmentations during training by either selecting augmented samples according to their loss or by weighting the loss contributions of each augmented sample using a trainable weight vector that is trained simultaneously with the parameters of the neural network. In a second step, we propose incorporating a unique hyperparameter to regulate the strength of the augmentations by grouping all hyperparameters for each augmentation. 

The contributions of this paper are summarized as follows:
\begin{itemize}
    \item We define two sample-adaptive augmentation policies to weight augmentation methods and propose the use of a single distortion parameter to optimize magnitude of the transformation.
    \item Using these two adaptive policies we demonstrate improvement in financial time-series results and state-of-the-art results in time-series datasets from the UCR archive \cite{UCR2018}. 
\end{itemize}

\section{Related Work}
While individual data augmentation methods are ubiquitous in neural networks regularization and to improve model robustness, developing automated data augmentation to find optimal strategies for combining different transformations has garnered a lot of interest in recent years. 
An early attempt to combine transformations is TANDA (Transformation Adversarial Networks for Data Augmentations) that uses a generative adversarial approach. In the first stage, a generative sequence model over user-specific transformations is learned (Generator), and then fed to a discriminator \cite{tanda}. Finally, the trained sequence generator is used to augment the training set for an end discriminative model. They test their model on three benchmark datasets for image (CIFAR-10 and MNIST), text (Automatic Content Extraction, ACE) and Digital Database for Screening Mammography and show improvements in all three. 
In a similar fashion, AutoAugment \cite{autoaugment} automatically searches for improved data augmentation policies in a reinforcement learning framework by using a controller that chooses an augmentation policy from the search space, then a secondary network (end model) is trained and its validation accuracy is used as reward using a policy gradient method to update the controller. Using an adversarial approach, Adversarial AutoAugment jointly optimizes the end model and augmentation policy search, with the controller attempting to increase the training loss of the end model thus, expecting the model to be robust against difficult examples. The model is tested on image benchmark datasets (CIFAR-10, CIFAR-100, SVHN and ImageNet) and shows an improvement in accuracy on all benchmarks.

A much simpler approach is RandAugment \cite{randaugment}, that for each mini-batch during training, randomly samples with equal probability $\frac{1}{K}$ a set of transformations from a group of $K$ available data augmentation methods and applies them to the mini-batch. It has two parameters to consider, the number of successive transformations applied on the mini-batch $N$ (normally one to three) and the magnitude of each augmentation distortion $M$ (how much an image is transformed), therefore, it has a significantly reduced search space which allows to use a simple grid search to optimize the parameters. It matches or surpasses all previous augmentation approaches on CIFAR, SVHN and ImageNet benchmarks. 

All the described methods so far learn an unchanging policy for all training samples. In contrast, our proposed methods learn a sample-adaptive policy that learns to weight the contribution of the individual transformations or selects the most relevant ones based on the loss of the sample. Furthermore, our approach is the first method to implement and automated data augmentation policy on time series data. 

Wu~\etal \cite{uncertaintyaug} propose an uncertainty-based random sampling scheme where all samples of a mini-batch are augmented by multiple augmentation methods randomly selected by a list of possible choices. Then all augmented samples are introduced to the network to calculate their corresponding losses. The augmented samples with the highest losses are used for updating the parameters of the network, meaning that some of the original samples do not contribute to the parameters update at all. In contrast, we follow a sample-adaptive augmentation strategy; we apply all augmentations to each sample and automatically adapt the contribution of each augmentation and of the original sample or select a subset of augmentations based on their losses. We observed that by keeping the augmented samples with the highest loss has a negative effect when using an ensemble of augmented samples to contribute to the update of the network parameters.

\section{Augmentation policies}
\label{sec:aug}

We propose two sample-adaptive augmentation policies that allow the use of multiple augmentation methods simultaneously by either learning a weight that multiplies the loss contributions of each augmentation, or selecting a subset of augmented samples by ranking their predicted loss. 

\begin{figure}
    \centering
    \includegraphics[width=\linewidth]{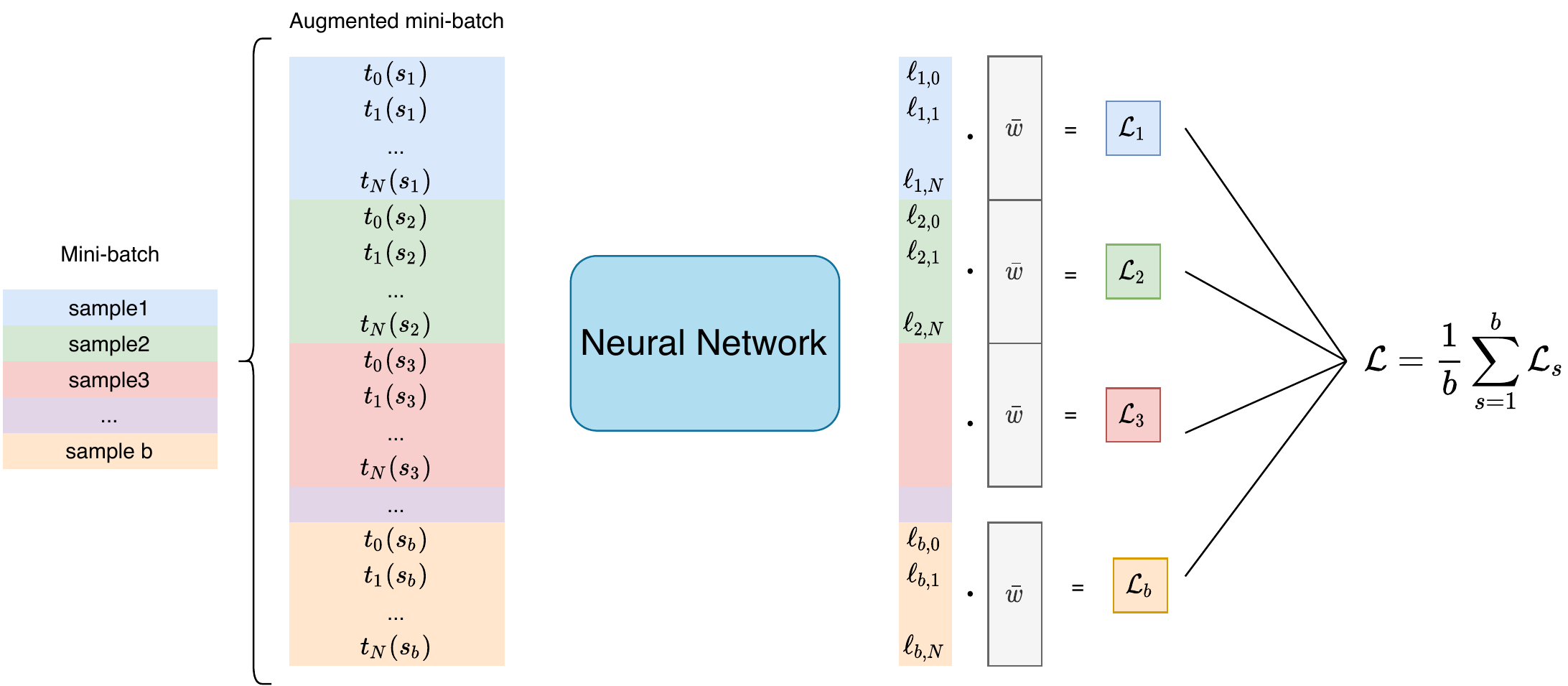}
    \caption{The proposed W-Augment policy applies all $N$ transformations on each sample on the mini-batch, and then each subgroup of samples is multiplied by the normalized weight $\bar{w}$ obtaining a scalar loss for each group of samples. Finally, the losses are averaged and a scalar loss $\mathcal{L}$ for the mini-batch is obtained and used to train the network.}
    \label{fig:adap_weight_b}
\end{figure}

\begin{figure*}
    \centering
    \includegraphics[width=\textwidth]{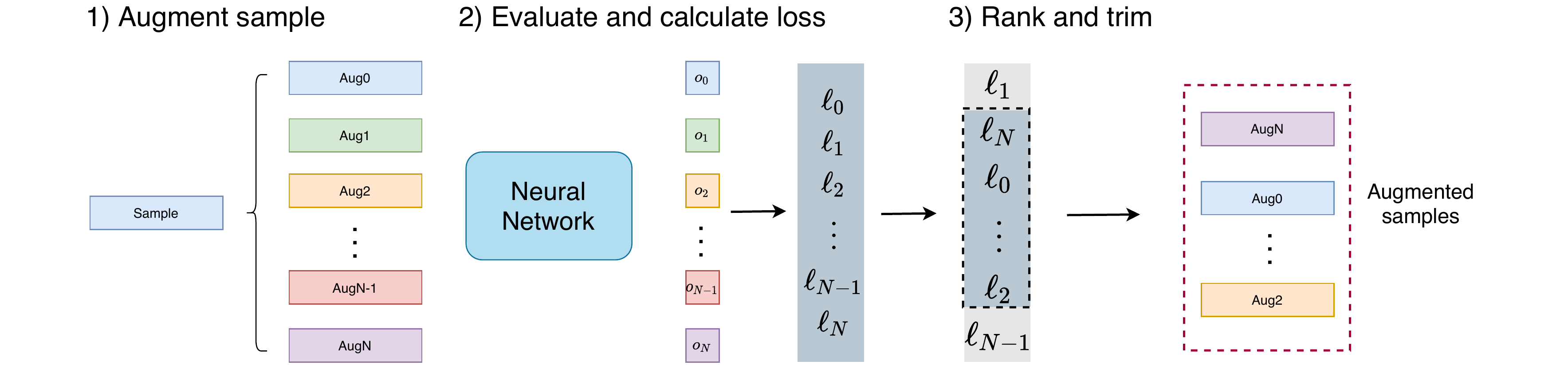}
    \caption{The proposed $\alpha$-trimmed Augment policy augments the sample with all available $N$ transformations, and evaluates the samples in the neural network, computing each individual loss. The losses are ranked and the top/bottom $\alpha$ samples that correspond to the top/bottom $\alpha$ losses are discarded. The subset of augmented samples are then used for training. Here a value of $\alpha=1$ is used.}
    \label{fig:alpha_trim}
\end{figure*}

The first proposed method, W-Augment, is shown in figure \ref{fig:adap_weight_b}. 
Given a sample $x_i$ with $i=1, \ldots, B$ where $B$ is the mini-batch size, the method applies all $N$ augmentations on each sample, obtaining $x_{i,j}$ with $j=0,\ldots,N$ where $j=0$ corresponds to the original sample without augmentation. Then, samples $x_{i,j}$ are evaluated in the neural network, computing the cross-entropy loss but without aggregating all individual losses (\i.e. normally the loss is averaged over all samples in a mini-batch and then backpropagated). The losses from the sample and the augmented transformations are multiplied using dot product by a trainable vector $\boldsymbol{\omega}$ of $N+1$ dimensions. In order to ensure all elements of the vector are positive and add up to one, we apply a softmax function $\sigma(\boldsymbol{w})$ to its values before the multiplication with the loss, i.e.:
\begin{equation}
  \mathcal{L}_i = \ell_{i,j} \cdot \sigma(\omega_j),
\end{equation}
where $\ell_{i,j}$ corresponds to the cross-entropy loss of sample $i$ augmented with transformation $j$. Each weighted loss $\mathcal{L}_i$ is then averaged over all elements on the mini-batch to calculate the loss that will be used by backpropagation to update the parameters of the network. The weight vector is initialized with constant number $1/(N+1)$. 

Our second proposed method, $\alpha$-trimmed Augment, works in two stages. Similarly to W-augment, each sample is augmented using all $N$ transformations, then evaluated in the neural network and each loss is computed separately. However, in $\alpha$-trimmed Augment we rank the losses and trim the $\alpha$ top and the $\alpha$ bottom ones, with $\alpha\in\mathbb{N}$; this means that we discard the augmented samples with the highest and lowest losses. 
The rationale behind this is that augmentations resulting in small losses will not lead to relevant training, meaning that such an augmented sample does not contribute to the learning process. While it can be argued that the highest losses should be preferred, as some methods propose; however, this might mean the model tries to learn only the hardest to classify samples while worsening its performance on easier samples. Finally the selected augmented samples are used to train the network. Figure \ref{fig:alpha_trim} illustrates the steps of $\alpha$-trimmed Augment.

Both our proposed methods select or weight augmentations in a sample-adaptive automatic manner; further, we want to determine the optimal magnitude for each transformation. In a second step - by taking advantage of recent developments in data augmentation policies such as RandAugment \cite{randaugment} and Population Based Augmentation (PBA) \cite{PBA2019} that demonstrate that it is sufficient to use a single distortion magnitude for all transformations instead of searching over optimal magnitudes for each augmentation - we also define a single distortion magnitude $M$ and optimize this value for all transformations. That is, for each transformation we propose a range of valid hyper-parameters to select from and a single value $M$ that will be optimized for all parameters simultaneously. This reduces the search space of the problem dramatically, given that we only need to optimize $M$ for W-augment and $M$ and $\alpha$ for $\alpha$-trimmed Augment. This allows us to apply a simple grid search to find optimal values that outperform current state-of-the-art methods.

\section{Experiments}
\begin{table}
	\caption{Augmentation methods used in the S$\&$P500 dataset, with each parameter and their fixed value used.}
	\centering
	    \resizebox{\linewidth}{!}{
	\begin{tabular}{@{}clcc@{}}\toprule
		ID & Augment method & Parameter & Value \\
		\cmidrule{1-4} 
		0 & Identity & & \\
		1 & Magnify & $t_0$ & [50,150]\\
		2 & Convolve & window & hahn\\
		3 & Pool & size & 3\\
		4 & Jitter & $\sigma$ & 0.01 \\
		5 & Quantize & level & 25 \\
		6 & Time Warp & knots, $\sigma$ & 4, 0.2 \\
		7 & Magnitude Warp & knots, $\sigma$ & 4, 0.2 \\
		8 & Window Warp & Window ratio, window scales & 0.1, $\{0.5, 2\}$ \\
		9 & Scaling & $\sigma$ & 0.1 \\
		10 & Reverse & ---  & --- \\
		\bottomrule
	\end{tabular}}
	\label{table:aug_sp}
\end{table} 
\subsection{Comparison method}
\begin{table*}
    \caption{Performance of the $k = 10$ long-short portfolios after transaction costs for the LSTM model and S$\&$P500 dataset.}
    \centering
    \small
    \resizebox{\linewidth}{!}{
\begin{tabular}{llccccccccccc}
\toprule
{}                      &     & Avg ret & Ann ret & Ann vol & IR &  D. Risk & DIR & Acc & F1\\
\midrule
{\bf A} & None                                &   0.13  &   34.64 &   28.43 &  1.22 &  18.78 &   1.84 &  51.03$\pm$0.97 &   48.51$\pm$2.05  \\
& W-augment          &   0.17  &   46.04 &   30.45 &  1.51 &  20.33 &   2.26 &  51.06$\pm$1.03 &  48.79$\pm$2.14   \\
& $\alpha_t$-Augment $(\alpha=1)$               &   {\bf 0.19}  &    {\bf 54.5} &   31.02 &  {\bf 1.76}  &  20.22 &    {\bf 2.7} &   51.1$\pm$1.01 &  48.97$\pm$2.39 \\
& $\alpha_t$-Augment $(\alpha=2)$               &   0.16  &   44.69 &   29.22 &  1.53  &  19.17 &   2.33 &  51.12$\pm$0.96 &  49.18$\pm$1.99  \\
& RandAugment      &   0.17  &   48.76 &   29.14 &  1.67  &  18.98 &   2.57 &  51.04$\pm$0.99 &   48.41$\pm$2.69  \\
\midrule
{\bf B} & W-augment $(+2)$        &   0.17  &   46.34 &   30.44 &  1.52 &  20.51 &   2.26 &    51.1$\pm$1.0 &   49.13$\pm$2.14  \\
& W-augment $(+3)$        &   0.18  &   52.02 &    30.1 &  1.73 &  20.35 &   2.56  &   51.09$\pm$1.0 &    48.99$\pm$2.11\\
& W-augment $(+4)$        &   0.16  &   43.76 &   28.39 &  1.54 &  18.72 &   2.34   &   51.12$\pm$1.0 & 49.56$\pm$1.57 \\
& $\alpha_t$-Augment $(+3, \alpha=1)$ &   0.16  &   42.98 &   26.77 &  1.61   &  17.36 &   2.48 & 51.11$\pm$0.98 &     49.02$\pm$2.2 \\
\bottomrule
\end{tabular}
}
\label{table:ch7-port}
\end{table*}
For comparison, we implement RandAugment \cite{randaugment}, an automated data augmentation method that has not been used on time-series data before. We choose RandAugment because it doesn't require a separate search phase on a proxy task and is competitive or outperforms previous automated approaches \cite{autoaugment,fastautoaugment}. 
RandAugment was originally proposed for classification tasks in computer vision and uses a set of 14 augmentations that are standard in computer vision such as rotate, solarize, posterize etc. Also included in the set of transformations is the identity, which leaves the sample unchanged. As in our methods, RandAugment uses a single global distortion parameter $M$ that regulates the strength of the transformations and a second parameter $N$ that corresponds to the number of consecutive transformations to be applied to the data, with $N$ a value usually between 1 and 3. The algorithm selects for each batch a transformation with uniform probability $\frac{1}{14}$. Because the algorithm only has two parameters, the search space is extremely small, and the authors find that using a simple greed search is quite effective. 

In this work we use RandAugment where the $N$ parameter is always one, therefore, we do not apply successive transformations to the time-series. 
The reason for this is that in computer vision, it would make sense to apply successive transformations such as Autocontrast and rotate to an image, this would leave the label unchanged. But in time-series, applying successive transformations might change the pattern excessively which might lead to a change in class.   
Instead of using computer vision transformations, we use the set of time-series augmentations from Table \ref{table:aug_sp} for the S$\&$P500 dataset and the augmentations from Table \ref{table:aug_ucr} for the UCR datasets.

\subsection{Stock classification with S$\&$P500 dataset}

We use the daily returns of all the constituents stocks of the S$\&$P500 index, from $1990$ to $2018$. This is a large-scale, significant dataset in the finance domain, because is representative of the US stock market. 
We follow the pre-processing scheme proposed by \cite{Krauss2017} where the data is divided into splits of $1000$ days, with a sliding window of $250$ days, which means that each split overlaps with the previous one by 750 days. A model is trained on each period, resulting in 25 trained models, one on each split. The data is segmented into sequences consisting of $240$ time steps $\{\tilde{R}^{s}_{t-239}, \ldots, \tilde{R}^{s}_{t}\}$ for each stock $s$, with a sliding window of one day, we use the first 750 days (approximately 3 years) for training and the last 250 days (1 year) for testing. This results in a training set of approximately 225K samples ((750-240)*500) and a test set of approximately 125K samples.

The data is standardised by subtracting the mean of the training set ($\mu_{train}$) and dividing by the standard deviation ($\sigma_{train}$), i.e., $\tilde{R}^{s}_t = \frac{R^{s}_t - \mu_{train}}{\sigma_{train}}$, with $R^s_t$ the return of stock $s$ at time $t$. The problem is defined as a two-class classification task with label $1$ when the returns of stock $s$ at time $t$ are above the daily median ($Y^s_{t+1}=1$) or $0$ when the returns are below the daily median ($Y^s_{t+1}=0$).

\subsubsection{Implementation details}

The augmentation methods used for this dataset are shown in Table \ref{table:aug_sp}, along with the parameters that control the augmentation and their values. We used a fixed value of the parameters in all cases except {\it Magnify}, where the starting point of the magnifying window is sampled with uniform probability between a range of values [50, 150]. The proposed parameter values have been shown to work in previous studies \cite{fons2020da}, \cite{Iwana2020}. 

We use the network proposed by \cite{Fischer2018} which consists of a single layer LSTM with $25$ neurons, and a fully connected two-neuron output. We use a learning rate of $0.001$, batch size $128$ and early stopping with patience $10$ with RMSProp as optimizer.

In order to evaluate the data augmentation methods in a task-specific setting, we build a simple trading rule in the following way: stocks are ranked daily by their predicted probability of belonging to a class (up or down trend), we then take the top $10$ and bottom $10$ stocks and build a long-short portfolio by equally weighting the stocks. Portfolios are analysed after transaction costs of 5 basis points (bps) per trade, where 1bps=$0.01\%$.  

\subsubsection{Results}

We test the proposed policies using methods 0 to 6 on the Table \ref{table:aug_sp}. Previous work indicates that these augmentation methods work well in stock datasets \cite{fons2020da}.
We are interested in evaluating the augmentation policies in a financial setting, where the focus is on portfolio performance instead of only classification accuracy; hence we evaluate the proposed trading rule using Information Ratio (IR) \cite{IR}, the ratio between excess return (portfolio returns minus benchmark returns) and tracking error (standard deviation of excess returns). Since the portfolios are long-short, they are market-neutral, thus there is no need to subtract a benchmark. We also include the average daily return (Avr ret) in percentage and calculate the downside information ratio (DIR), the ratio between excess return and the downside risk (variability of underperformance below the benchmark), which differentiates harmful volatility from the overall volatility.

Panel A from Table \ref{table:ch7-port} presents the results for W-augment, $\alpha$-trimmed Augment and RandAugment. We can see that all three methods achieve a higher return than not using an augmentation policy, with a volatility and downside risk similar to the baseline. 

Beyond the transformations that have been shown to work on this financial dataset, we study adding more transformations to our policies, which correspond to augmentations 7 through 10 from Table \ref{table:aug_sp}. These results are shown in panel B of Table \ref{table:ch7-port} where W-Augment $(+2)$ corresponds to augmentations 0 to 6 plus 7 and 8 ({\it Magnitude Warp} and {\it Window Warp}), W-Augment $(+3)$ includes the previous transformations as well as transformation 9, and so on. We can see that all automatic augmentation policies outperform the baseline (no transformation) and W-Augment benefits in all cases from the additional transformations. 
\begin{figure*}
    \centering
    \includegraphics[width=0.24\textwidth]{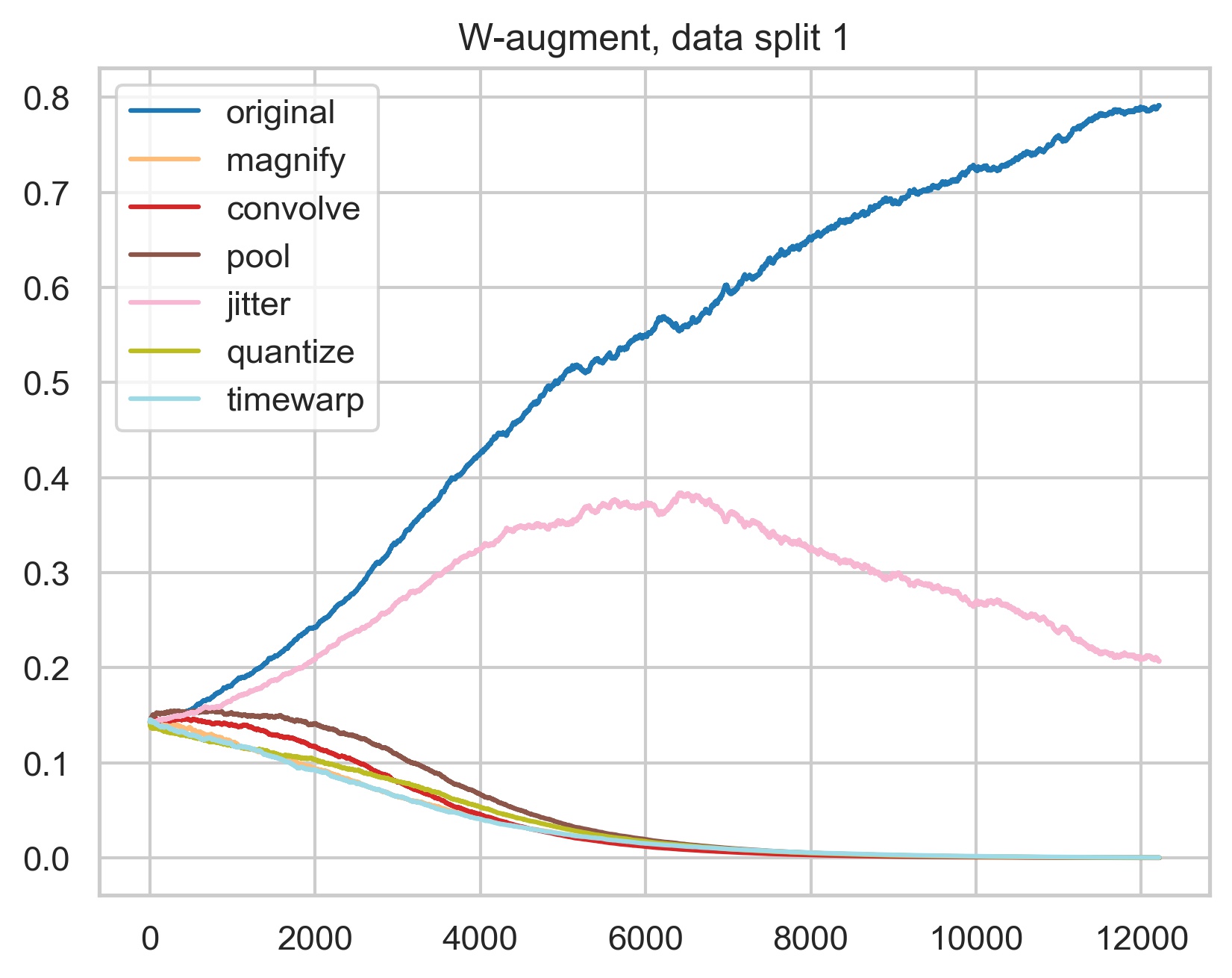}~
    \includegraphics[width=0.24\textwidth]{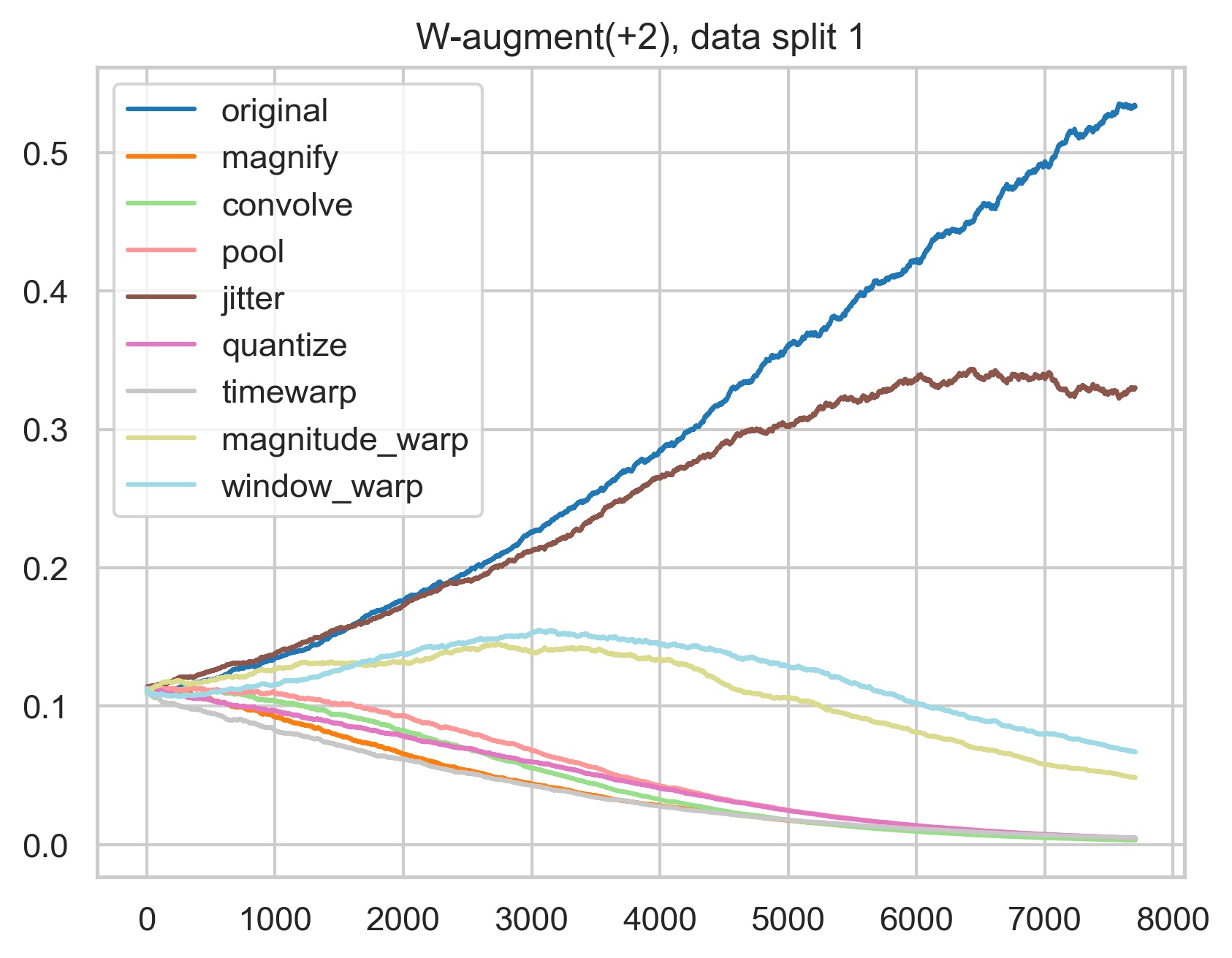}~
    \includegraphics[width=0.24\textwidth]{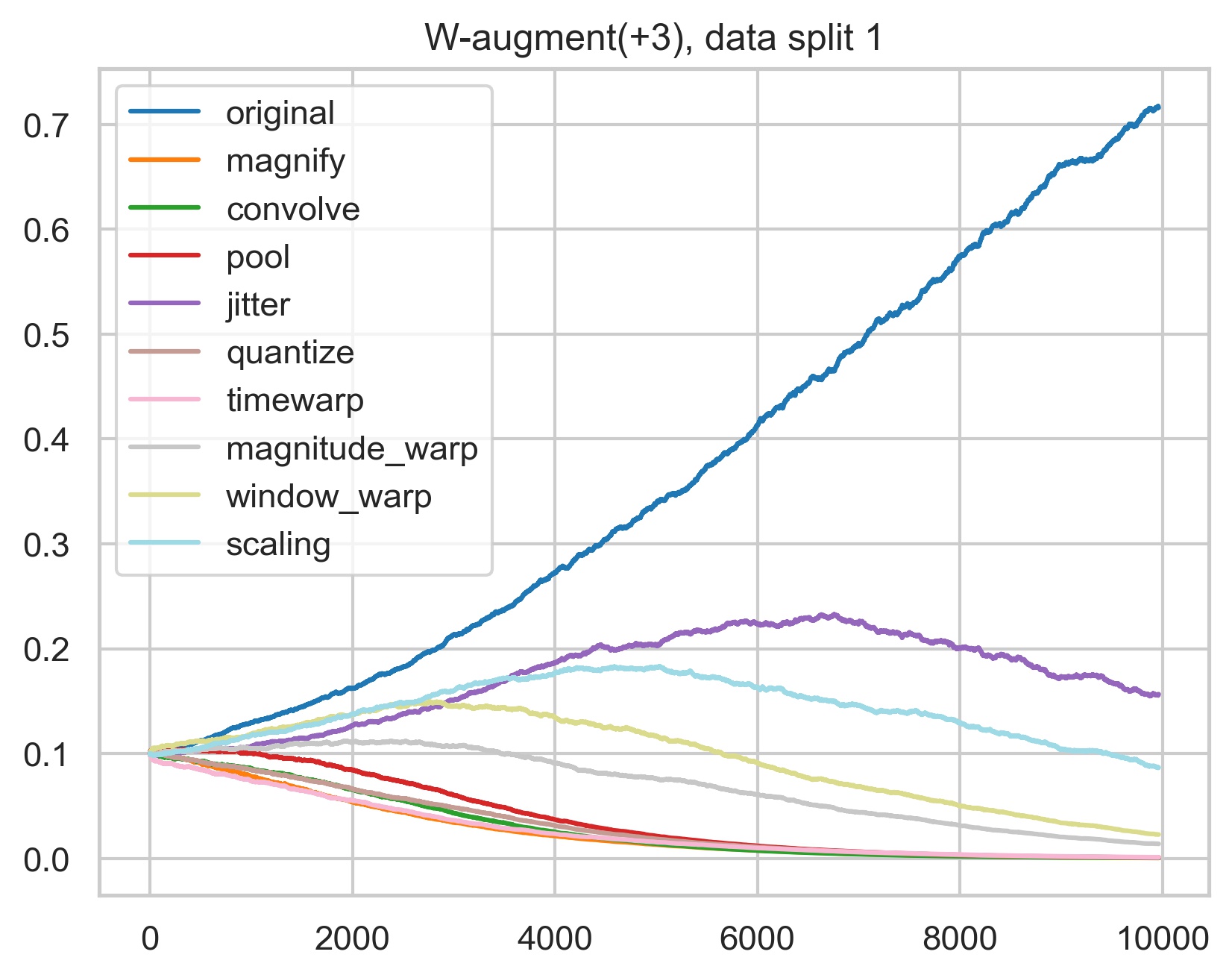}~
    \includegraphics[width=0.24\textwidth]{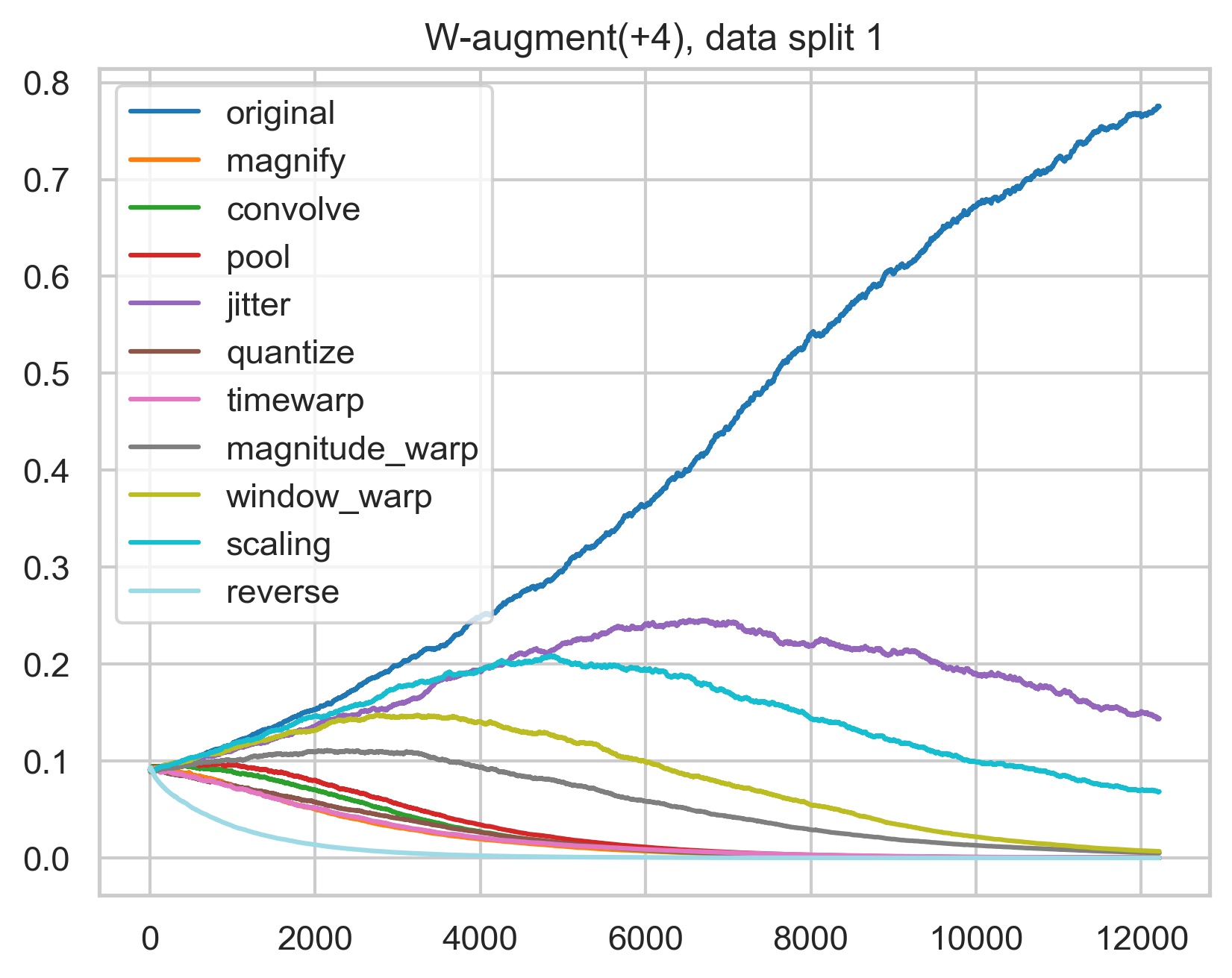}
    \caption{Weights assigned to each augmented sample per training mini-batch. Each column corresponds to W-augment, W-augment $(+2)$, W-augment $(+3)$ and W-augment $(+4)$.}
    \label{fig:weights_waug_finance}
\end{figure*}~
\begin{figure*}
    \centering
    \includegraphics[width=0.25\linewidth]{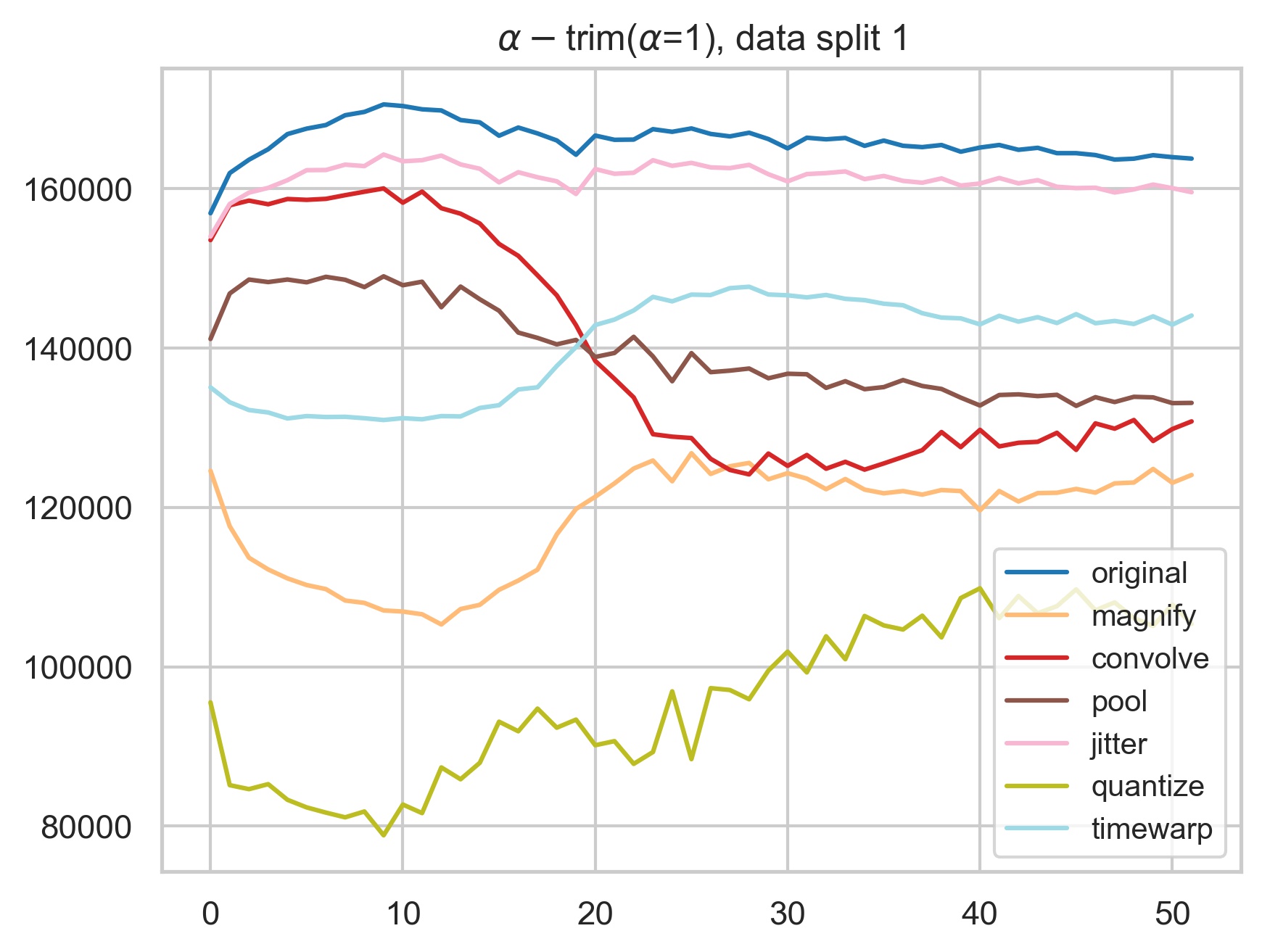}~\includegraphics[width=0.25\linewidth]{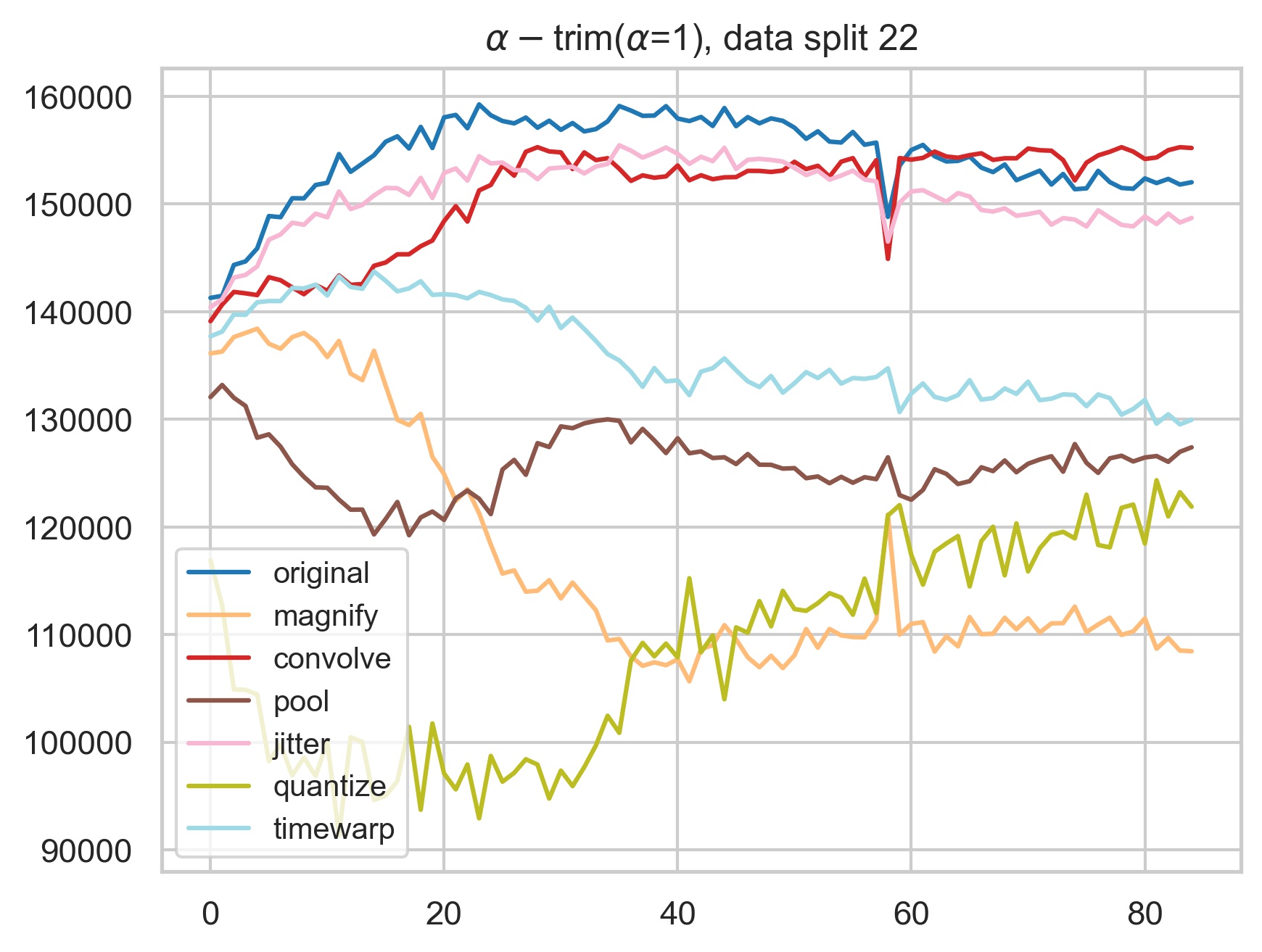}~
    \includegraphics[width=0.25\linewidth]{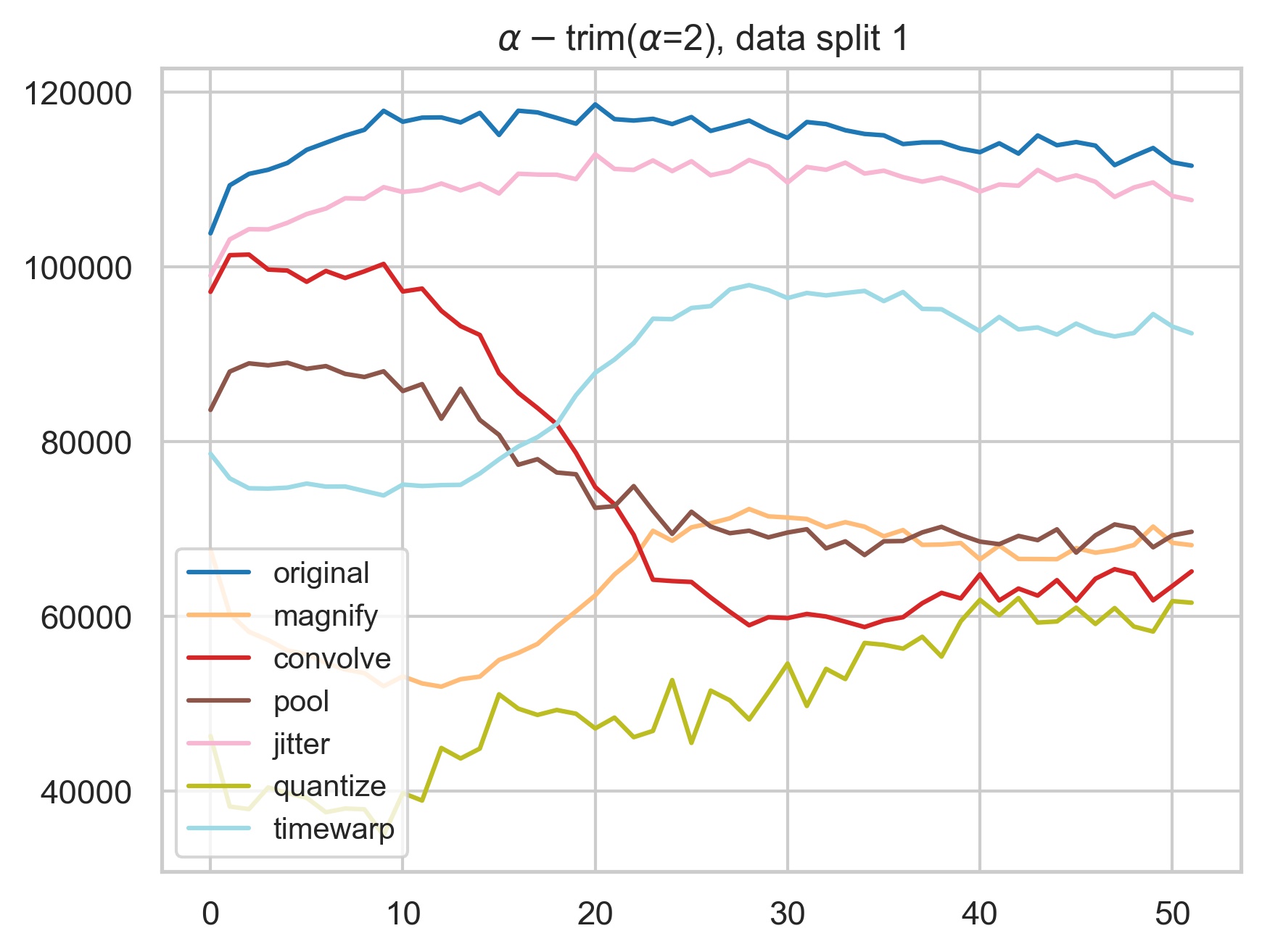}~\includegraphics[width=0.25\linewidth]{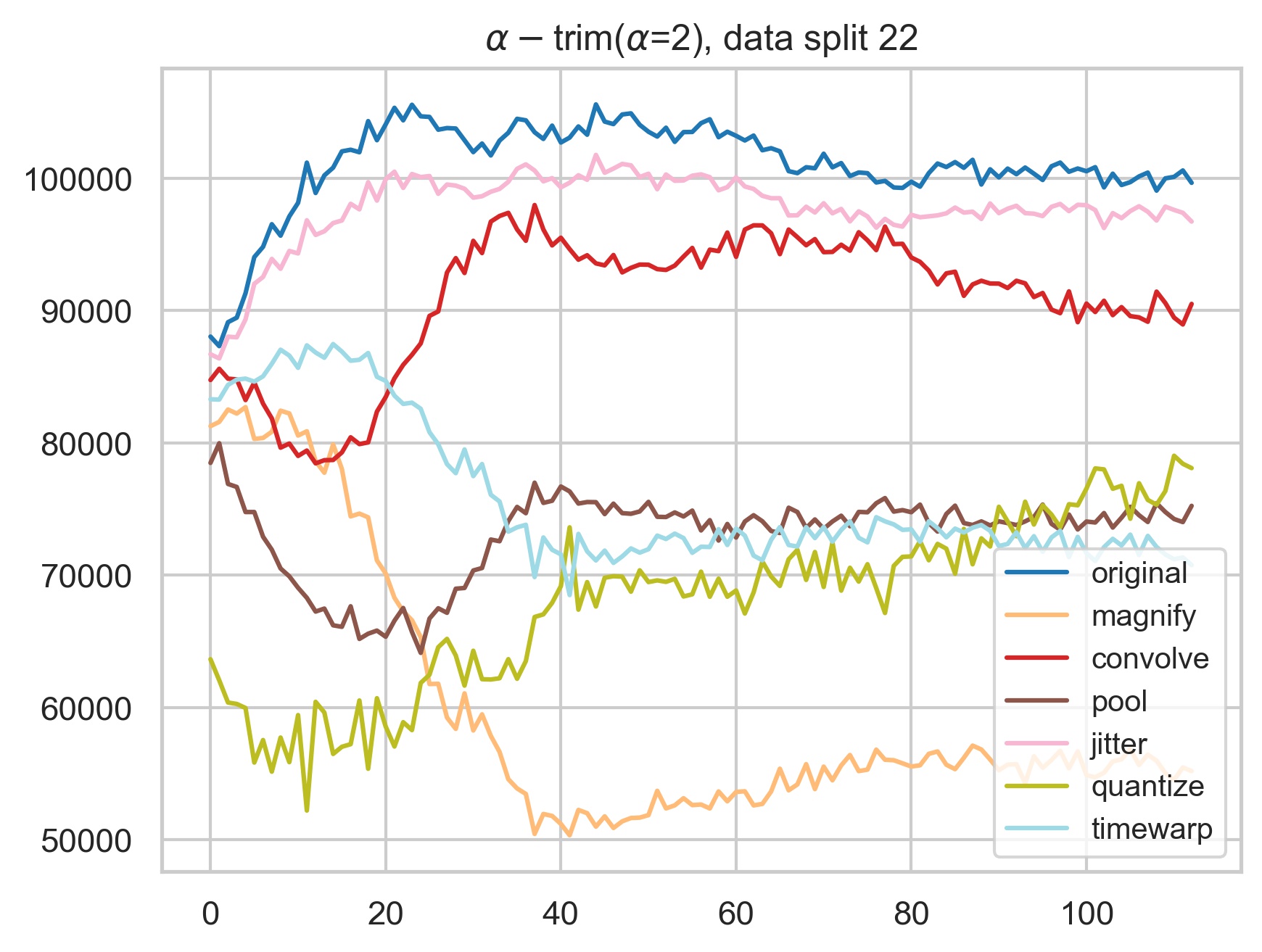}
    \caption{Number of times the augmentation method was selected per epoch on $\alpha$-trimmed Augment.}
    \label{fig:histos_atrim_finance}
\end{figure*}

Figure \ref{fig:weights_waug_finance} shows the learned weights assigned to each augmentation method during training, with each column corresponding to W-Augment with the original transformations and W-Augment with the additional ones, all for the same split of data. 
We can see that in all cases, the method tends to put more weight in the original sample over time. In general, the weighting is quite consistent even when considering an additional number of augmentations \eg jitter gets a higher weighting than most of the other augmentation methods in all cases.
Interestingly, in the case of \eg W-Augment $(+2)$ (on the bottom plot) where {\it Reverse} is included, the transformation is very quickly assigned a zero weight. This is an augmentation method that we know is not useful in financial time-series data. 

Figure \ref{fig:histos_atrim_finance} corresponds to the $\alpha$-trimmed Augment policy and shows the number of times each augmentation method was selected per batch. We can see that, as with W-Augment, the original sample gets selected most frequently, and along with jitter, both remain quite constant, whilst other methods change over time.
We can see that for both values of $\alpha$, the behaviour remains the same, \ie the frequency with which each transformation gets selected is in a similar proportion \wrt the value of alpha. We do observe a change in behaviour between splits of data. We show data splits 1 and 22, therefore there are around 21 years of difference between both datastes. We can see that {\it convolve} is the fifth method in frequency on the last epochs on split 1, while it becomes the most frequently selected on split 22. This suggests that not all augmentation methods are equally effective in all time periods, especially given that financial time-series are non-stationary, and $\alpha$-trimmed Augment might be more sensitive to this phenomenon. The plots for both policies on all data splits can be found in the supplementary material.  

\subsection{UCR datasets}
\begin{table}
	\caption{Augmentation methods used on the UCR dataset, with their tunable parameters and their range of values. The distortion magnitude $M$ is a linear interpolation on each range.}
	\centering
    \resizebox{\linewidth}{!}{
	\begin{tabular}{@{}clcc@{}}\toprule
		ID & Augment method & Tunable parameters & Range \\
		\cmidrule{1-4} 
		0 & Identity & & \\
		1 & Jitter & $\sigma$ & [0.01, 0.5] \\
		2 & Time Warp & knots, $\sigma$ & $\{3,4,5\}$, [0.01, 0.5] \\
		3 & Window slice & ratio  & [0.95, 0.6] \\
		4 & Window Warp & Window ratio, window scales & 0.1, [0.1, 2] \\
		5 & Scaling & $\sigma$ & [0.1,2.0] \\
		6 & Magnitude Warp & knots, $\sigma$ & $\{3,4,5\}$, [0.1, 2] \\
		7 & Permutation & Max segments & $\{3,4,5,6\}$ \\
		8 & Dropout & $p$ & [0.05, 0.5] \\
		\bottomrule
	\end{tabular}
	}
	\label{table:aug_ucr}
\end{table} 

 \begin{table*}
    \caption{\label{table:ucr_4}
Test accuracy $(\%)$ on 15 datasets from UCR archive. Comparisons across the default training without augmentation policy (baseline) and W-Aug, $\alpha$-trimmed Augment and RandAugment all using 4 transformations in addition to the identity. $M_W$, $M_{\alpha}$ and $M_R$ are the optimal distortion magnitude for W-Aug, $\alpha-$trim and RandAugment, respectively.}
\begin{subtable}{\linewidth}
    \centering
    \small
\begin{tabular}{lccccccc}
\toprule
{}                            &  Baseline  &  W-Augment(4) &  $M_W$  &  $\alpha_t$-Augment(4) &  $M_{\alpha}$ &  RandAugment(4) &  $M_R$ \\
\midrule
ECG5000                       &     94.5 &       94.2 & 15  &        93.8 &  5 & \textbf{94.6} & 1    \\
EthanolLevel                  &     85.9 &       \textbf{86.6} & 5 &        84.6 &  5 &  86.0 & 1  \\
ProximalPhalanxOutlineCorrect &     91.7 &       91.5 & 10   &        \textbf{92.1} &  20 & 92.1 & 5  \\
MiddlePhalanxOutlineCorrect   &     82.9 &       84.7 &  5 &        83.2 &  20  & \textbf{86.3} & 5 \\
DistalPhalanxOutlineCorrect   &     \textbf{78.0} &       76.9 &  15  &        77.2 &  5  & 77.5 & 5  \\
Strawberry                    &     97.5 &       97.2 &  10 &        \textbf{97.6} &  1 &  97.0 & 5  \\
MixedShapesSmallTrain         &     \textbf{91.5} &       89.6 &  20   &        90.9 &  5 & 90.3 & 1   \\
InlineSkate                   &     \textbf{51.6} &       36.1 &  5  &        41.6 &  1 &  46.0 & 1  \\
ECG200                        &     86.0 &       89.0 &  1   &        \textbf{90.0} &  1 & 87.0 & 15  \\
ACSF1                         &     \textbf{91.0} &       88.8 &  5 &        86.0 &  5 &  88.0 & 1   \\
Ham                           &     66.7 &       80.6 &  15   &        \textbf{81.0} &  20  & 75.2 & 15   \\
Haptics                       &     \textbf{57.5} &       55.5 &  10    &        54.5 &  1   &  48.4 & 15  \\
Fish                          &     98.3 &       \textbf{99.0} &  15  &        98.3 &  5 &  98.3 & 1  \\
WormsTwoClass                 &     77.9 &       78.7 &  10   &        79.2 &  5  & \textbf{80.5} & 1  \\
Worms                         &     \textbf{83.1} &       81.0 &  1   &        80.5 &  1 &  \textbf{83.1} & 10  \\
\bottomrule
\end{tabular}
\end{subtable}
\vspace{2em}
    \caption{\label{table:ucr_8} Test accuracy $(\%)$ on 15 datasets from UCR archive. Comparisons across the default training without augmentation policy (baseline) and W-Aug, $\alpha$-trimmed Augment and RandAugment all using 4 transformations in addition to the identity. $M_W$, $M_{\alpha}$ and $M_R$ are the optimal distortion magnitude for W-Aug, $\alpha$-trimmed Augment and RandAugment, respectively. }
\begin{subtable}{\linewidth}
    \centering
    \small
\begin{tabular}{lccccccc}
\toprule
{}                            &  Baseline  &  W-Augment(8) &  $M_W$  &  $\alpha_t$-Augment(8) &  $M_{\alpha}$ &  RandAugment(8) &  $M_R$ \\
\midrule
ECG5000                       &     \textbf{94.5} &            94.2 & 5   &        \textbf{94.5} &   1  &        94.4 &  1   \\
EthanolLevel                  &     \textbf{85.9} &         84.0 &  10   &        81.4 &   5   &          83.4 &  1   \\
ProximalPhalanxOutlineCorrect &     91.7 &       92.4   & 5  &        \textbf{92.8} &   15    &           91.1 &  20      \\
MiddlePhalanxOutlineCorrect   &     82.9 &       82.7 &   1   &        83.2 &   5     &         \textbf{84.5}  &  5  \\
DistalPhalanxOutlineCorrect   &     \textbf{78.0} &       75.6 & 5 &         75.4 &  10     &       76.8 &   1   \\
Strawberry                    &     97.5 &       97.6 &  15 &         \textbf{97.8} &   5   &       \textbf{97.8} &    1   \\
MixedShapesSmallTrain         &     91.5 &       \textbf{92.2} &  10  &        91.6 &    15    &         91.3 &  5 \\
InlineSkate                   &     \textbf{51.6} &       38.1 & 5  &        43.3 &     1  &           38.2 &   1 \\
ECG200                        &     86.0 &       \textbf{91.0} & 1   &        89.0 &     1     &          88.0 &   15  \\
ACSF1                         &     \textbf{91.0} &        89.0 & 20   &         90.0 &    10  &           90.0 &   1  \\
Ham                           &     66.7 &        \textbf{77.0} &  20  &         75.2 &  10     &          76.2 &   10      \\
Haptics                       &     \textbf{57.5} &       52.9 &  1     &        54.9 &   5    &          53.9 &  15      \\
Fish                          &     98.3 &        \textbf{99.4} &  20   &     \textbf{99.4} &      10       &           97.7 &   1   \\
WormsTwoClass                 &     77.9 &              \textbf{79.5} &  10    &        79.2 &    5     &          79.2 & 10   \\
Worms                         &     \textbf{83.1} &         80.0 &  10  &        79.2 &    1    &       80.5 &  1  \\
\bottomrule
\end{tabular}

\end{subtable}
\end{table*}

In order to test our proposed sample-adaptive augmentation methods on time-series classification problems from a different domain, and show that it can be used to improve generalization in tasks beyond financial prediction. We use a subset of datasets from the UCR archive \cite{UCR2018}, which consists of 128 univariate time-series datasets of various types and characteristics. Given the large amount of data in the UCR archive, we select a subset of datasets based on training size, focusing on small datasets (with training samples between 100 and 200 samples), and medium datasets that have around 500 training samples. 
The secondary reason to evaluate our augmentation policies in these datasets is that the UCR archive is a well established baseline for time-series classification tasks, with state-of-the-art results being updated regularly. 

For each dataset, the samples $x_s$ are normalized by subtracting the mean of the sample and dividing by the standard deviation $\tilde{x}_s = \frac{x_s - \mu_s}{\sigma_s}$. In order to fine tune the hyperparameter $M$, we separate the training set into training and validation with a proportion of 80/20$\%$. Because in some cases the datasets are quite small and have multiple classes, we do the train/validation split in a stratified way in order to preserve the class proportion in both sets.

\subsubsection{Implementation details}

The augmentation methods used on these datasets are shown in Table \ref{table:aug_ucr} where each method contains one or two parameters that control the augmentation magnitude and the range that the parameters can take. 
In this case, we will optimize the distortion magnitude $M$ to find a collective value of distortion that maximizes performance on the cross-validation set.

Following \cite{fawazInception} we use the InceptionTime network - currently the only state-of-the-art model in the UCR archive based on deep neural networks.  
Inception time is an ensemble model that trains 5 instances of the inception network with different initializations and averages the prediction scores of the five models before outputting the predicted label. The original work was trained without validation, using all the training set data, for 1500 epochs and Adam optimizer with an initial learning rate of 0.001. Learning rate was reduced based on the plateau of the training loss (with patience 50 and factor 0.5) and the model with best training accuracy was saved. We slightly modify the training methodology by training over 5 stratified shuffled splits, and use the validation loss to reduce the learning rate. In order to speed-up training, we used early stopping with a patience of 150 epochs and a maximum number of epochs of 1500. Everything else was the same as used in InceptionTime.

\subsubsection{Results}

We first implemented the three policies, W-Augment, $\alpha$-trimmed Augment and RandAugment using four augmentations and the identity (transformations 0 to 4 from Table \ref{table:aug_ucr}). $M$ can take values from 1 to 20 and we selected from the set $\{1,5,10,15,20\}$. Given that InceptionTime is an ensemble of five models, we trained each policy with a value of $M$ on InceptionTime and averaged the performance of the validation set over the five splits of data. For each dataset we selected the $M$ with the highest validation accuracy. For comparison, we trained InceptionTime with the same methodology but without using a augmentation policy. Table \ref{table:ucr_4} shows the accuracy of the best performing policy on validation set, and the corresponding value of $M$ for the dataset. We can see that W-Aug(4) beats or equals the baseline on 6 of the test times \wrt the baseline, $\alpha$-trimmed Augment 7 times and RandAugment wins or draws 10 times, resulting in overall better performance in accuracy. In general, the values selected by W-Augment tend to be higher, which might account for the lagging in performance \wrt RandAugment. 

Table \ref{table:ucr_8} shows the accuracy of the best performing policies on the validation set when using all eight transformations from Table \ref{table:aug_ucr}. We see in this case that W-Aug(8) and $\alpha$-trimmed(8) Augment win or draw 8 and 9 times respectively \wrt the baseline, and RandAugment(8) only wins 5 times. This could mean that the extra augmentations might hurt performance, as RandAugment selects each transformation with equal probability, it cannot mitigate sub-optimal augmentations. 
On the other hand, the proposed W-Augment and $\alpha-$trimmed Augment, by weighting or selecting the contribution of each augmentation in a sample-adaptive automatic manner, can successfully suppress such sub-optimal augmentations.

\begin{figure}
    \centering
    \includegraphics[width=0.5\linewidth]{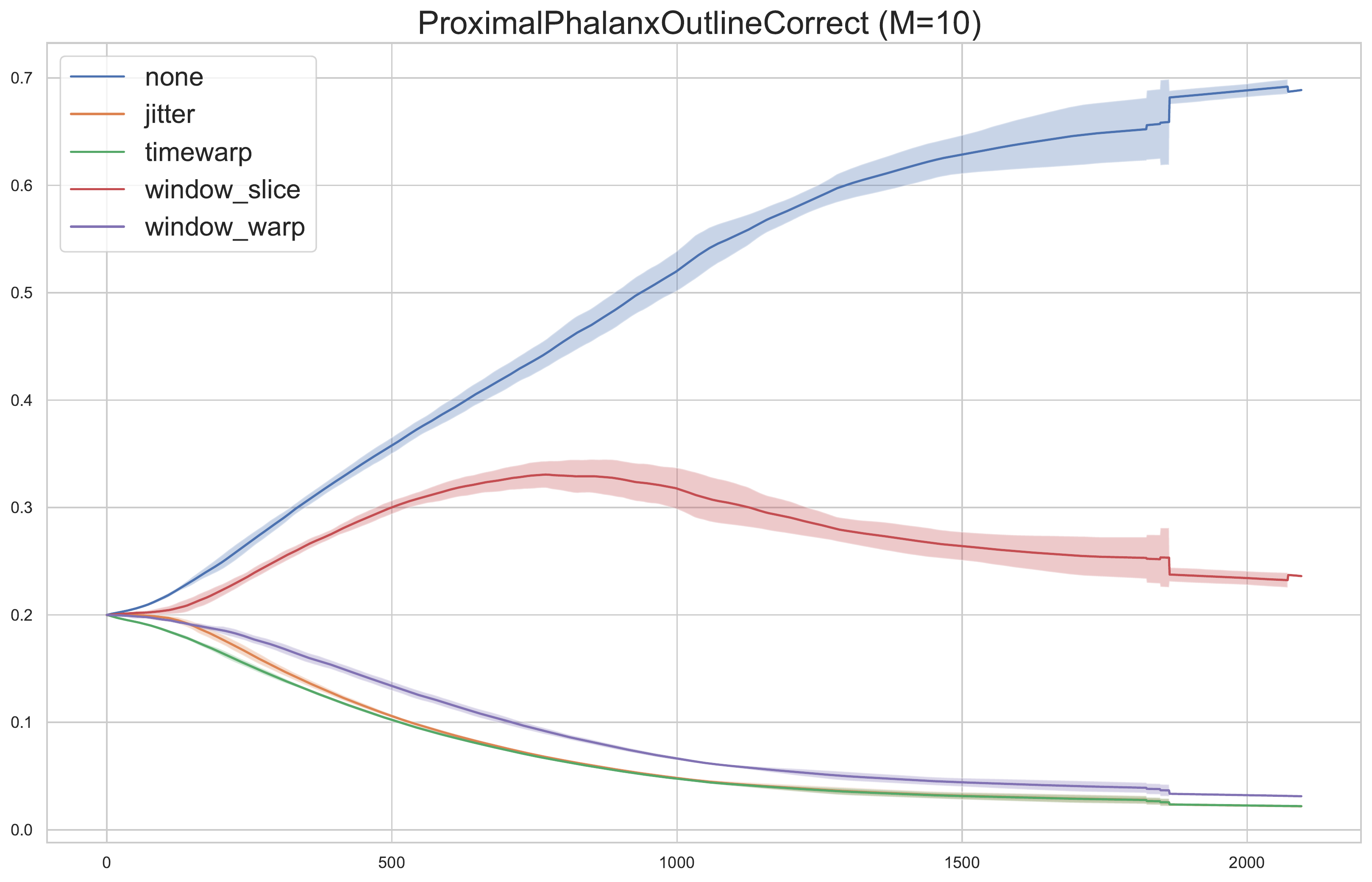}~\includegraphics[width=0.5\linewidth]{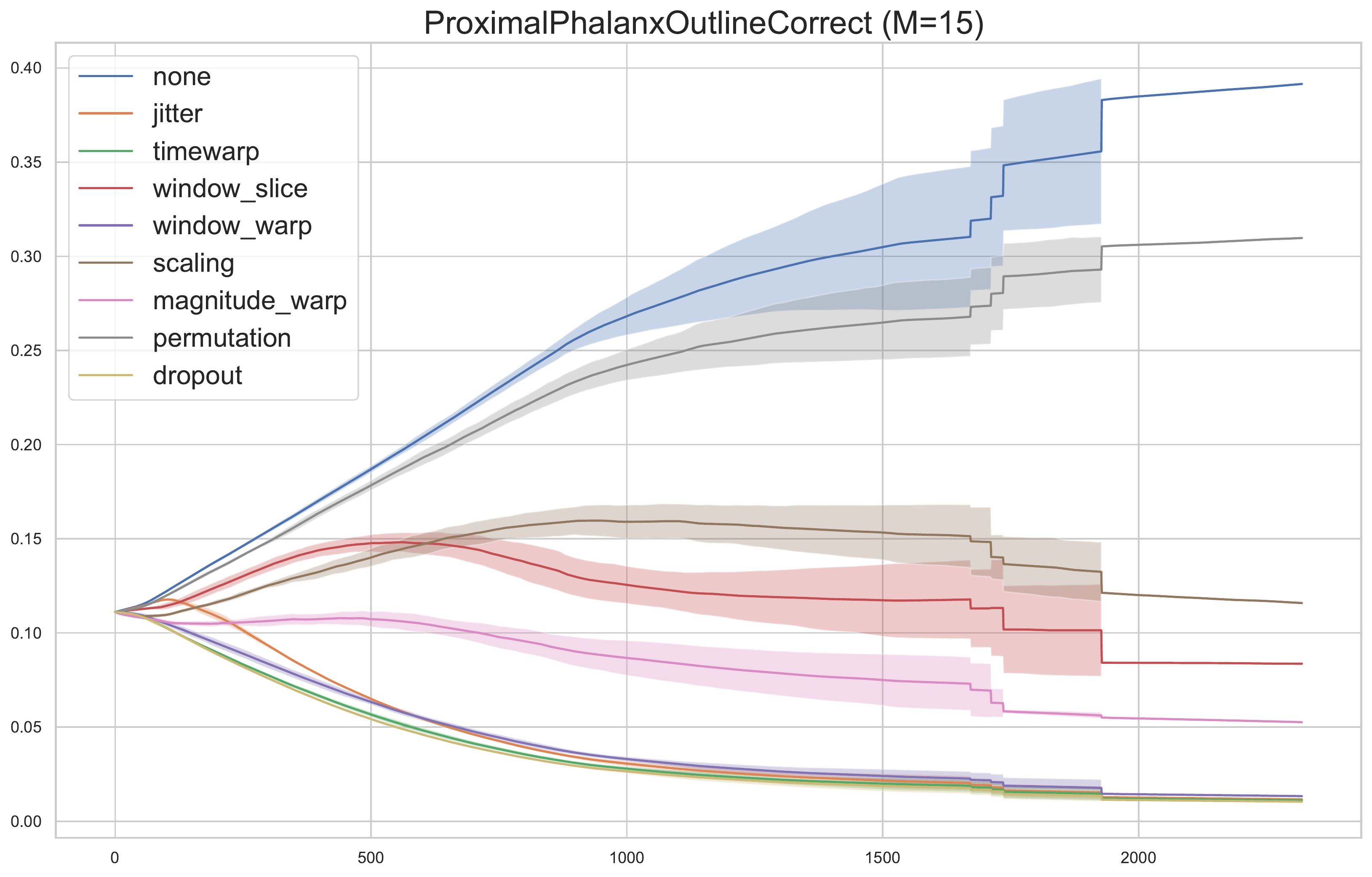}\\
    \includegraphics[width=0.5\linewidth]{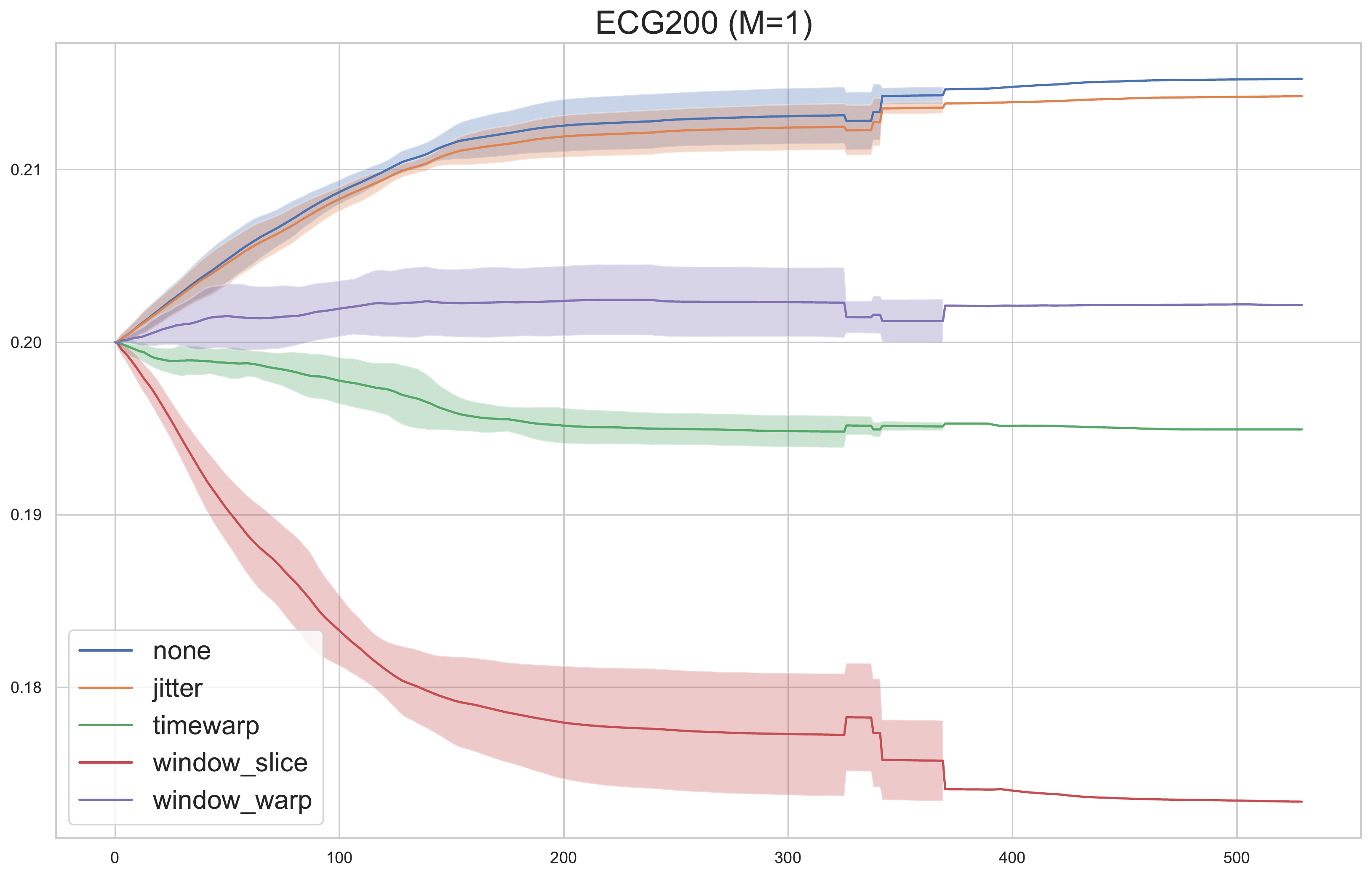}~\includegraphics[width=0.5\linewidth]{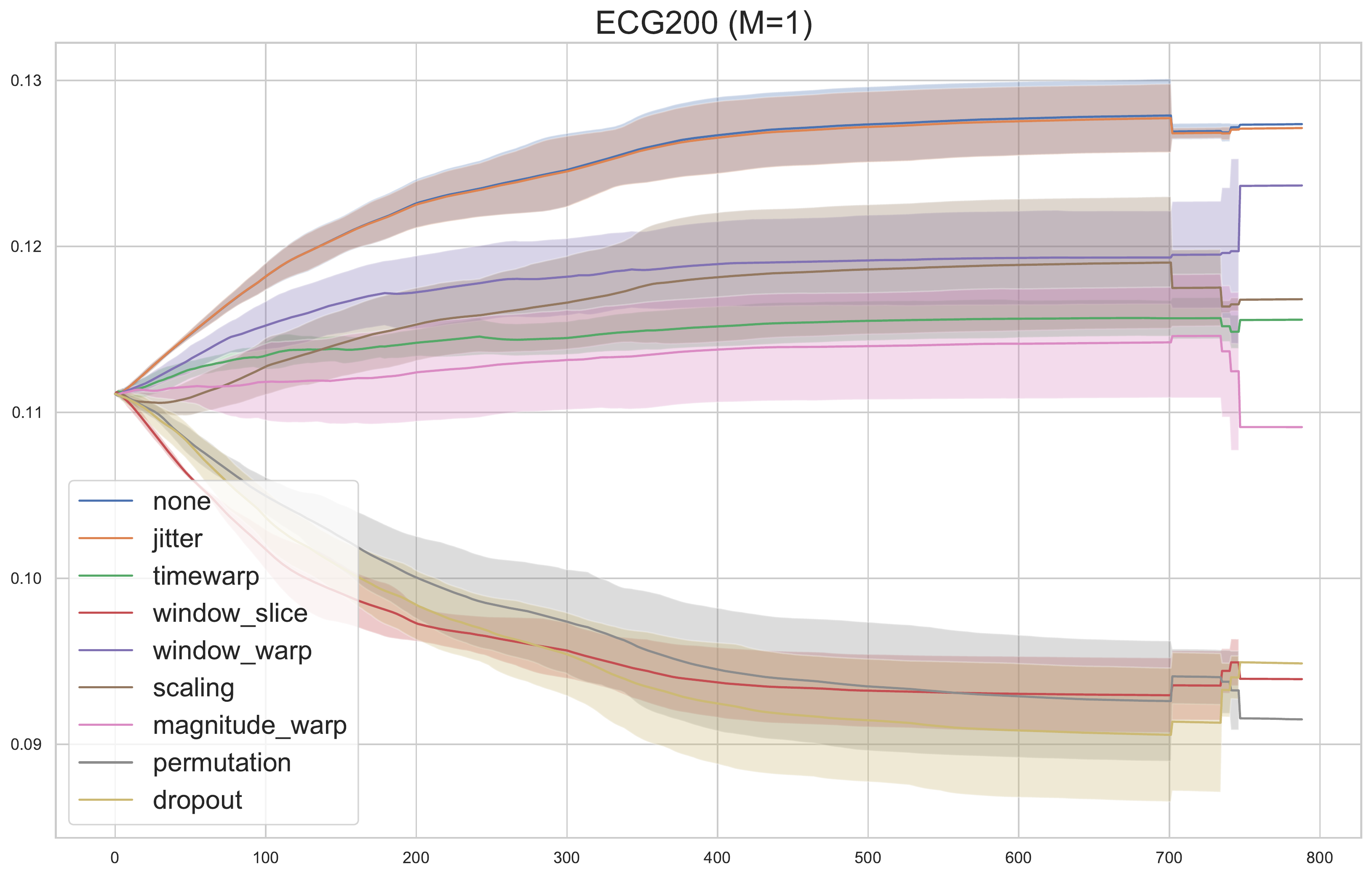}
    \caption{Weights assigned to each augmented sample on the mini-batch per training iteration. Plots on the left correspond to W-Augment trained  using 4 transformations and plots on the right corresponds to W-augment trained with 8 augmentations.}
    \label{fig:weights_ucr}
\end{figure}

Figure \ref{fig:weights_ucr} shows the learned weights assigned to each method during training on four datasets for the optimal value of $M$. Given that InceptionTime consists of 5 trained networks that are ensembled, we have 5 policies in place per model. Therefore, the plots show the average of the weight per training iteration and the shaded part corresponds to plus/minus one standard deviation. We see that there is agreement on the weights in all models. 
In all cases, W-Augment assigns a higher weight to the original sample, but it changes behaviour with different types of datasets. For example, on the ProximalPhalanxOutlineCorrect dataset it learns a weight of zero for {\it time warp}, but this method is more relevant on the ECG200 dataset. 

\begin{figure}
    \centering
    \includegraphics[width=0.85\linewidth]{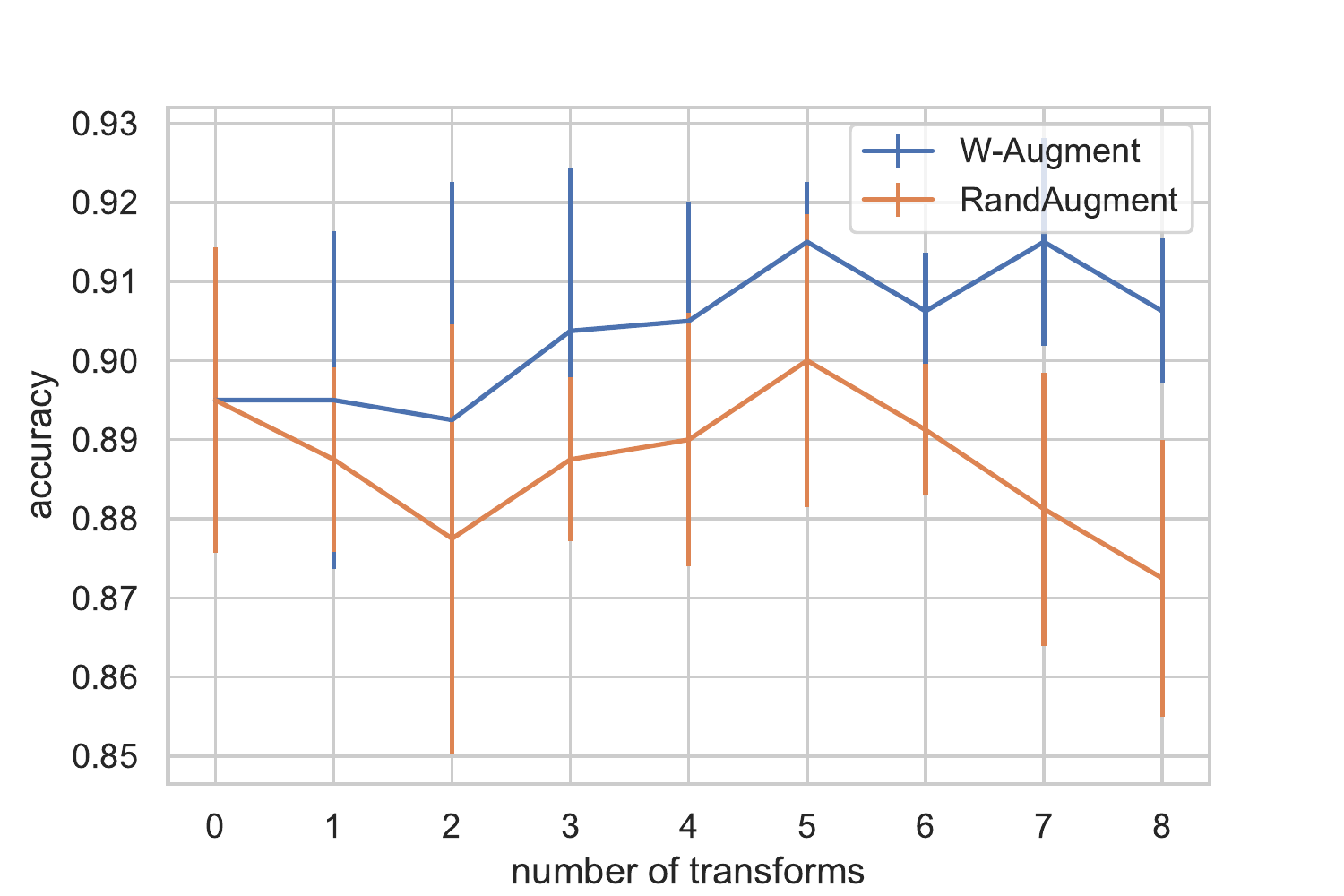}\\
    \includegraphics[width=0.85\linewidth]{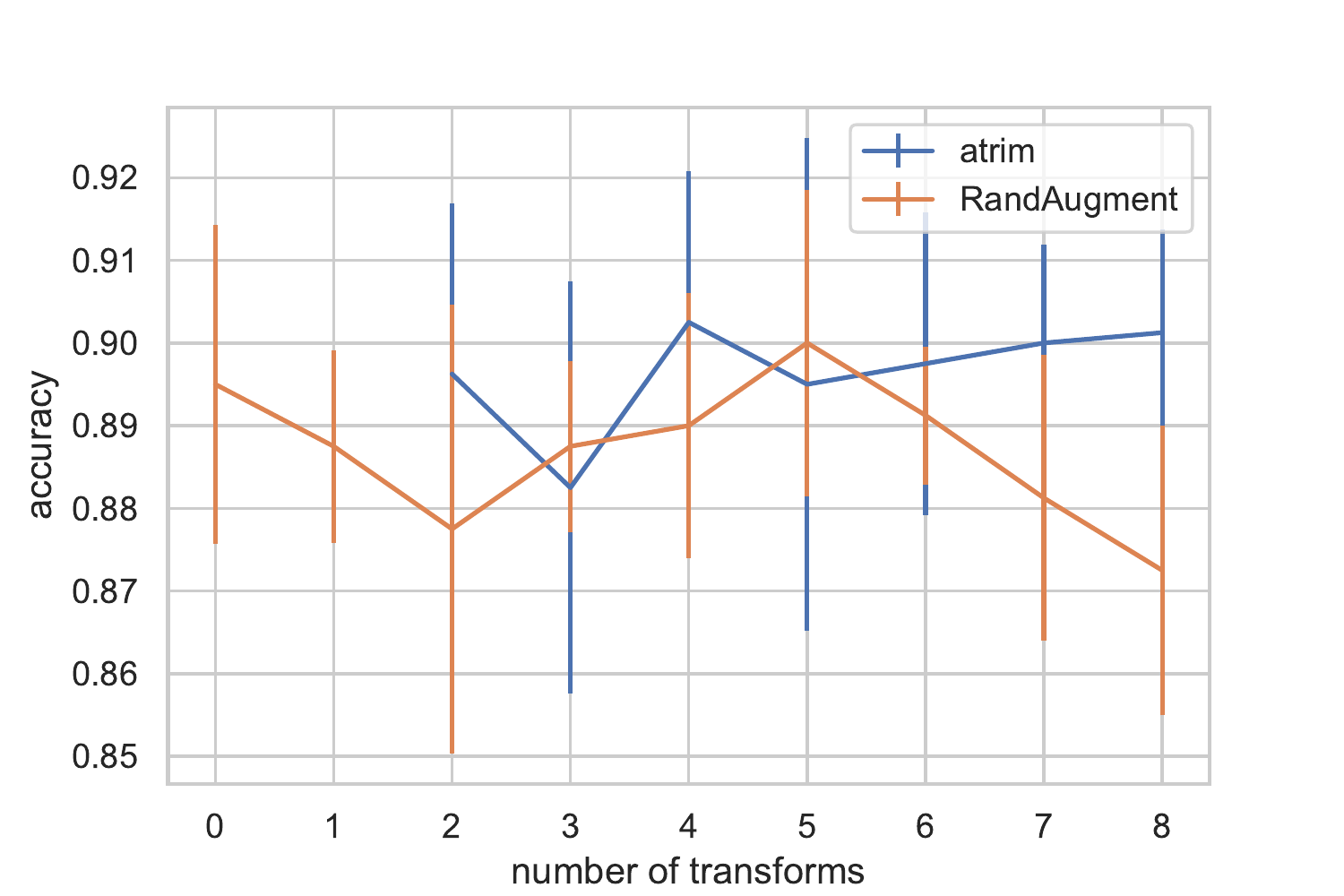}
    \caption{Mean validation accuracy for InceptionTime on the ECG200 dataset trained with W-Augment($M=1$), $\alpha$-trimmed Augment($M=1$) and RandAugment($M=15$) using randomly sampled subsets of transformations. Error bars indicate one standard deviation.}
    \label{fig:number_of_transforms}
\end{figure}

Finally, we study the dependence of the included transformations. Figure \ref{fig:number_of_transforms} shows the mean validation accuracy on the ECG200 dataset for all three methods using randomly sample subsets of the list of 8 transformations from Table \ref{table:aug_ucr}. We can see that W-Augment tends to improve with more augmentation methods, as well as $\alpha$-trimmed Augment.

\section{Conclusions}
\label{sec:ch7conc}

In this paper, we have presented two sample-adaptive automatic weighting schemes for data augmentation: W-Augment learns to weight the contribution of the augmented samples to the loss and $\alpha$-trimmed Augment selects a subset of transformations based on the ranking of the predicted training loss. We have validated our proposed policies on the S$\&$P500 dataset and on datasets from the UCR archive. 
For comparison we implemented RandAugment and tested it on time-series data for the first time, achieving competitive results when using a small number of transformations. 

On the real-world financial dataset, we showed that the augmentation methods in combination with a trading strategy lead to improvements in annualized returns of over 50$\%$, and on time-series data coming from other domains we outperform state-of-the-art models on over half of the datasets, and achieve similar performance in accuracy on the remainder.

In this work we have focused on univariate time-series. An interesting further problem is to test the proposed adaptive augmentation methods on multivariate time-series.  
Additionally, future work should study how the proposed augmentation methods apply to other machine learning domains, where data augmentation has been proven to increase generalization, such as image, speech and audio recognition. 
We believe that both our methods should work well on large datasets without incurring costly computational costs given that it adds at most two hyperparameters.

\bibliography{biblio}
\bibliographystyle{icml2021}

\end{document}